%% file: main.tex
\documentclass[10pt,twocolumn,letterpaper]{article}

\usepackage{wacv}              

\input{preamble}

\usepackage{adjustbox}
\usepackage{float}

\definecolor{wacvblue}{rgb}{0.21,0.49,0.74}
\usepackage[pagebackref,breaklinks,colorlinks,allcolors=wacvblue]{hyperref}

\def\wacvPaperID{00007} 
\def\confName{WACV}
\def\confYear{2026}

\title{SynSacc: A Blender-to-V2E Pipeline for Synthetic Neuromorphic Eye-Movement Data and Sim-to-Real Spiking Model Training}

\author{
Khadija Iddrisu$^{1}$* \quad
Waseem Shariff$^{2}$ \quad
Suzanne Little$^{1}$ \quad
Noel O'Connor$^{1}$ 
\\
$^{1}$Dublin City University, Dublin, Ireland 
$^{2}$University of Galway, Galway, Ireland 
\\
{\tt\small
khadija.iddrisu2@mail.dcu.ie} \\
{\tt\small
waseem.shariff@universityofgalway.ie \quad
\{suzanne.little, noel.oconnor\}@dcu.ie
}
}

\begin{document}
\maketitle
\input{sec/0_abstract}    
\input{sec/1_intro}

\input{sec/2_background}

\input{sec/3_dataset}

\input{sec/4_methodology}

\input{sec/5_experimentation}

\input{sec/6_results}
{
    \small
    \bibliographystyle{ieeenat_fullname}
    \bibliography{main}
}

\end{document}

%% file: preamble.tex
\usepackage{algorithm}
\usepackage{algpseudocode}
\usepackage{gensymb}
\usepackage{microtype}

\usepackage[none]{hyphenat}


%% file: sec/0_abstract.tex
\begin{abstract}
The study of eye movements, particularly saccades and fixations, are fundamental to understanding the mechanisms of human cognition and perception. Accurate classification of these movements requires sensing technologies capable of capturing rapid dynamics without distortion. Event cameras, also known as Dynamic Vision Sensors (DVS), provide asynchronous recordings of changes in light intensity, thereby eliminating motion blur inherent in conventional frame-based cameras and offering superior temporal resolution and data efficiency.
In this study, we introduce a synthetic dataset generated with Blender to simulate saccades and fixations under controlled conditions. Leveraging Spiking Neural Networks (SNNs), we evaluate its robustness by training two architectures and finetuning on real event data. The proposed models achieve up to 0.83 accuracy and maintain consistent performance across varying temporal resolutions, demonstrating stability in eye movement classification. Moreover, the use of SNNs with synthetic event streams  yields substantial computational efficiency gains over artificial neural network (ANN) counterparts, underscoring the utility of synthetic data augmentation in advancing event-based vision. All code and datasets associated with this work is available at \url{https://github.com/Ikhadija-5/SynSacc-Dataset}.
\end{abstract}

%% file: sec/1_intro.tex
\section{Introduction}


Visual perception in  biological and artificial systems is shaped by  eye movements such as fixations and saccades. \emph{Fixations} reveal the regions of interest where information is extracted, while \emph{saccades} reflect the dynamic shifts in visual attention and the efficiency of neural control systems. Together, they serve as critical indicators in domains such as reading, visual search, motor coordination, and decision-making.  Their significance have been examined in domains, including human – computer interaction~\cite{majaranta2019eye}, gaze-based interface design~\cite{chen2024identifying}, attention modeling~\cite{fischer1987mechanisms,remington1980attention}, etc. 

\begin{figure*}[!htbp]
    \centering
    \includegraphics[width=15cm,height=7cm]{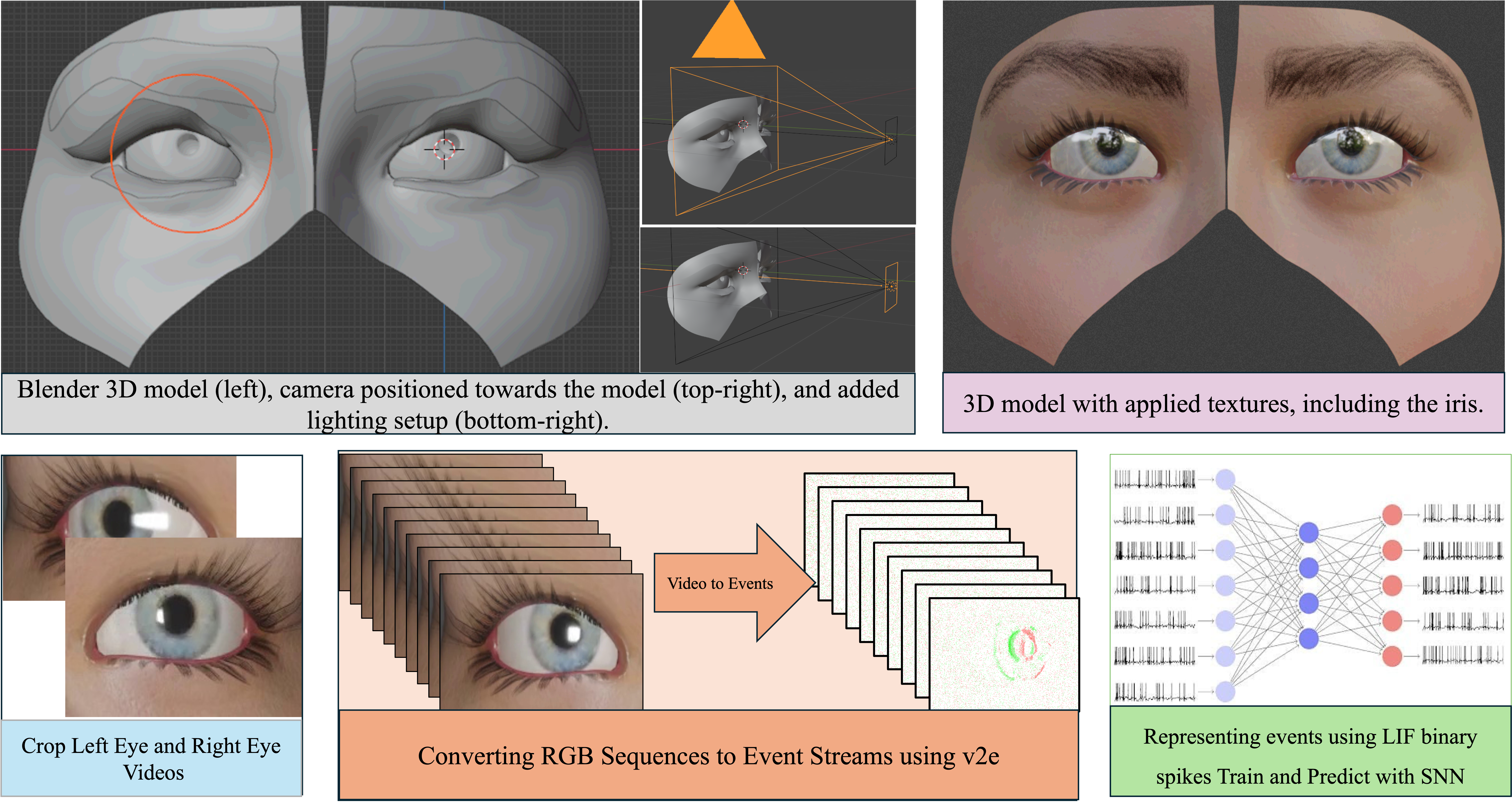}
    \caption{Overview of the data generation and processing pipeline: The workflow starts with a Blender 3D model, where camera and lighting settings are configured for realistic rendering. Textures, including detailed iris textures, are then applied to the model. The eyes are cropped from the rendered images to create focused sequences, which are subsequently converted into videos. These RGB sequences are transformed into event-based streams using the v2e framework. Finally, the event streams are represented as LIF binary spikes, which are used to train and predict with a spiking neural network (SNN).}
    \label{fig:workflow_overview}
\end{figure*}

Early work on fixation and saccade classification relied on highly invasive techniques such as scleral search coils~\cite{madariaga2023safide}, which provided millisecond-level accuracy but  limited by participant discomfort and restricted applicability outside laboratory settings. Signal-based eye trackers subsequently became the dominant modality, offering non-invasive measurement  with sufficient temporal resolution yet often constrained by calibration drift and reduced robustness in naturalistic environments~\cite{singh2012human}. Parallel efforts explored  EEG to capture neural correlates of oculomotor dynamics, but these approaches remained indirect and required complex signal interpretation~\cite{nikolaev2016combining,shigapova2024electrophysiology}. More recently, frame-based video analysis has emerged as a promising alternative, enabling fixation and saccade detection through direct observation of ocular movements in visual recordings~\cite{chang2021high}.  However, these methods struggle with latency, noise sensitivity, and the inability to capture micro-movements (motion blur) or high-speed transitions with sufficient granularity~\cite{iddrisu2024event}.

Event Cameras (ECs) are a distinctive alternative to conventional modalities. Unlike traditional cameras that capture full frames at fixed intervals, ECs operate asynchronously, recording only changes in pixel intensity. This mechanism produces a sparse stream of spatiotemporal data with microsecond temporal resolution. Fixations typically occur within defined temporal window  of 50-600ms, while saccades unfold over shorter, rapid intervals of 20-300ms reaching velocities as high as 700$\degree$~\cite{leigh2008using}. Conventional frame-based sensors often fail to capture these dynamics accurately due to limitations such as motion blur and redundant frame data. In contrast, the asynchronous data acquisition of ECs enables precise capture of both saccadic movements and fixation plateaus, thereby enhancing temporal precision while reducing computational overhead.

Spiking Neural Networks (SNNs) operate on discrete spikes rather than continuous activations and mimic the temporal dynamics of biological neurons. Their ability to integrate temporal information over time windows allows for robust classification of patterns directly from event streams. Moreover, the  sparsity of SNN activations, combined with event-driven computation, enables significant reductions in synaptic operations compared to traditional Artificial Neural Networks (ANNs) offering substantial gains in energy efficiency. While event camera data and the associated models built on SNNs offer great potential for less intrusive capture and more fine-grained analysis of eye movements, there is currently a lack of sufficiently annotated training data and, consequently, established models for classification of eye movement.

Capturing extensive, varied and detailed eye movement data from human subjects is expensive and time-consuming. Therefore to address this limited availability of event-based data and to further develop models for fine-grained movement classification, this study introduces a method that combines synthetic data with real EC data and trained using SNNs. Specifically, Blender is employed to construct controlled scenarios of saccades and fixations, yielding well-defined annotations as illustrated in ~\Cref{fig:workflow_overview}.  The key contributions of this paper are as follows:

\begin{enumerate}

\item We propose a method for generating synthetic eye movement data using the Blender 3D creation suite and subsequently generate synthetic events from this using an event simulator. The dataset generated is manually annotated into sequences of fixations and saccades, addressing the scarcity of such data in event-based eye movement research.

\item  We implement and evaluate two Spiking Neural Network (SNN) classification frameworks that we trained and evaluated across varying temporal resolutions of the generated saccades and fixations. Results demonstrate that the proposed SNN architectures exhibit significant computational efficiency compared to equivalent Artificial Neural Networks (ANNs).

\item Finally,  we evaluate the robustness and real-world applicability of the synthetic dataset through  transfer learning involving finetuning the best-performing SNN model with a real-world event-based eye movements dataset. This validates the utility of our synthetic data for pre-training and generalization to real event camera data.

\end{enumerate}

%% file: sec/2_background.tex
\section{Background}

\subsection{Event-based Eye Movements Datasets}
The sparse and event-driven nature of Event cameras (ECs)  closely mimics the architecture of biological vision and this alignment makes them particularly well-suited for capturing rapid eye movements such as saccades.~\cite{gallego2020event}. While several event-based datasets have been introduced for eye movement analysis tasks such as eye-tracking~\cite{ryan2021real,bonazzi2024retina}, gaze estimation~\cite{zhao2024rgbe,10188686}, pupil segmentation~\cite{3et}, microsaccades identification~\cite{shariff2025benchmarking}, etc., none of these have provided explicit ground truth  annotations for saccades and fixations~\cite{iddrisu2024event}. At present, the largest benchmark offering raw event streams for high-frequency eye movement analysis is the EV-Eye dataset~\cite{zhao2023ev}. EV-Eye is a large-scale, multimodal dataset collected from numerous participants, consisting of saccade and fixation states as well as smooth pursuit movements. However, while session-level movement types  may be indicated, the fine-grained, high-temporal-resolution labels of individual saccades and fixations, are  absent.


\subsection{Event-Driven Eye Movement Algorithms}
A number of studies have highlighted the potential of event-based data to capture rapid ocular movements with a high temporal precision~\cite{iddrisu2024event, chen2025event}. These include robust pupil localization and gaze tracking at kilohertz rates, enabling applications in augmented reality, human–computer interaction and generally in real time efficiency. Bonazzi et al.~\cite{bonazzi2024retina} propose a spiking regression model, Retina, which achieves a high precision with only 64k parameters and latency as low as 5.57ms. 
Jiang et al.~\cite{jiang2024eye} propose a SNN framework for efficient pupil localization while addressing high power consumption in frame-based systems. Similarly, Groenan et al.~\cite{groenen2025gazescrnn} propose GazeSCRNN for near-eye gaze tracking that incorporates a hybrid architecture for optimal spatio-temporal feature extraction resulting in a Mean Angle Error (MAE) of 6.034° and Mean Pupil Error (MPE) of 2.094mm. Furthermore, the 2025 Event-Based Eye Tracking Challenge summarized the latest algorithmic advances, with leading methods utilizing CNNs, Bi-GRUs, and attention mechanisms for robust pupil localization and gaze tracking~\cite{chen2025event}. However, despite progress the literature has primarily emphasized on coarse gaze estimation and overall eye motion analysis. No study to date has systematically addressed fine-grained movement such as classification of  saccades and fixations, leaving an important gap in event-based eye tracking research.

%% file: sec/3_dataset.tex
\section{SynSacc Dataset}
In this section, we describe the data generation pipeline, from the creation of Blender samples  to event  simulation.
\label{sec:dataset}

\begin{algorithm}
\caption{Generate Saccade-Based Eye Rotations}
\begin{algorithmic}[1]
\Require $A_{\text{left}}, A_{\text{right}}$ (armatures), $B_{\text{left}}, B_{\text{right}}$ (bones)
\Require $F_{\text{start}}, F_{\text{end}}, \Delta F, L, mirror\_y$
\Ensure $R_{\text{left}}, R_{\text{right}}$
\Function{GenerateSaccades}{$A,B,F_{\text{start}},F_{\text{end}},\Delta F,L$}
    \State $R \gets [\ ]$
    \For{$frame \gets F_{\text{start}}$ \textbf{to} $F_{\text{end}}$ \textbf{step} $\Delta F$}
        \State set Blender frame to $frame$
        \State $r \gets$ random rotation vector in $\pm L$ degrees
        \State apply Euler rotation $r$ to bone $B$
        \State insert keyframe
        \State append $r$ to $R$
    \EndFor
    \State \Return $R$
\EndFunction

\Function{ApplyMirroredRotations}{$A,B,F_{\text{start}},F_{\text{end}},\Delta F,R$}
    \State $idx \gets 0$
    \For{$frame \gets F_{\text{start}}$ \textbf{to} $F_{\text{end}}$ \textbf{step} $\Delta F$}
        \State $r \gets R[idx]$
        \State $r_{\text{mirror}} \gets (r_x, -r_y, r_z)$
        \State apply Euler rotation $r_{\text{mirror}}$ to bone $B$
        \State insert keyframe
        \State $idx \gets idx + 1$
    \EndFor
\EndFunction

\State $R_{\text{left}} \gets$ \Call{GenerateSaccades}{$A_{\text{left}},B_{\text{left}},F_{\text{start}},F_{\text{end}},\Delta F,L$}
\State $R_{\text{right}} \gets$ \Call{ApplyMirroredRotations}{$A_{\text{right}},B_{\text{right}},F_{\text{start}},F_{\text{end}},\Delta F,R_{\text{left}}$}
\State \Return $R_{\text{left}}, R_{\text{right}}$
\end{algorithmic}
\label{algo}
\end{algorithm}

\subsection{Blender Rendering}

SynSacc generates synthetic saccadic eye movements in a 3D Blender environment and converts them to event-streams using the V2E simulator~\cite{hu2021v2e}. The method operates on the armature bones of the left and right eyes, modifying their rotations frame by frame to simulate realistic eye motion. Users define the frame range, sampling interval, and a maximum rotation angle that bounds individual saccades. As shown in \Cref{algo}, the algorithm iterates over the specified frames for each eye. At every frame, a random 3D rotation vector within the defined limits is sampled and applied to the eye bone, with keyframes inserted to record the motion. All rotations are stored to preserve the temporal trajectory. This process is first applied to the left eye. To enforce coordinated binocular motion, the right eye trajectory is generated by mirroring the left eye rotations, inverting the Y-axis while retaining the X and Z components. The resulting rotations are applied at corresponding frames and keyframes, producing conjugate eye movements. The final output consists of synchronized rotation sequences for both eyes, passed through V2E to generate synthetic event streams. This fully synthetic pipeline enables scalable pretraining and fine-tuning of SNNs without collecting personal gaze data, ensuring GDPR-compliant neuromorphic research.

\subsection{Event Simulation}

The synthetic RGB eye movement sequences were converted into event data using the V2E (Video-to-Event) simulator, an open-source tool designed to generate realistic DVS events from frame-based counterparts~\cite{hu2021v2e}. The  process involves a Temporal upsampling step achieved through the Super-SloMo framework, an end-to-end convolutional neural network designed for multi-frame video interpolation~\cite{jiang2018super}. By generating intermediate frames with a slow-motion factor of $8$, the temporal resolution is substantially enhanced, thereby enabling more precise event generation. The upsampled, high-temporal-resolution frames are then processed by the event generation  model following the core principle of ECs. An event is triggered at a pixel whenever the change in logarithmic intensity $(ln(I))$ exceeds a predefined threshold.

\begin{figure}[!ht]
\centering
\begin{equation}
     E(x, y, t) = 
\begin{cases} 
+1, & \text{if } \Delta \ln I(x, y, t) \geq \theta_{ON}, \\
-1, & \text{if } \Delta \ln I(x, y, t) \leq -\theta_{OFF}, \\
0, & \text{otherwise},
\end{cases}
\end{equation}
\end{figure}

where $E(x,y,t)$ denotes the event at pixel $(x,y)$ and time $t$. Here, 
$\theta_{\mathrm{ON}} $and $\theta_{\mathrm{OFF}} $represent the positive and negative contrast thresholds, respectively, with $+1$ corresponding to ON events (brightness increase) and $-1$ to OFF events (brightness decrease).

The  event generation is configured using the ``Noisy'' preset model to include sensor artifacts, generating an output event stream matching the $346 \times 260$ resolution of the DVS346 sensor model~\cite{mueggler2017event}. Key parameters for the model includes an ``event threshold'' of $0.2$ (change in $ln(I)$); this value defines the sensor's sensitivity, requiring a $0.2$ change in logarithmic intensity $(ln(I))$ to trigger an event. This value balances event fidelity during subtle motion with robustness against low-level noise. A ``threshold standard deviation''  of $0.05$, a photoreceptor low-pass cutoff frequency of $30 Hz,$ a leak ``event rate'' of $0.1 Hz$, and a shot noise rate of $5 Hz$ is also applied. The chosen parameters (leak and shot noise) is deliberately included to introduce background activity and spatial noise variation. This ensures the resulting event data is robust and representative of a physical event sensor, preventing algorithms trained on the dataset from being overfit to perfectly clean event patterns. These configurations balance the need for high temporal resolution with the generation of a realistic, noisy event stream that accurately reflects a physical EC operating at a standard sensitivity level.

\begin{figure}[!htbp]
    \centering    \includegraphics[width=1.0\linewidth]{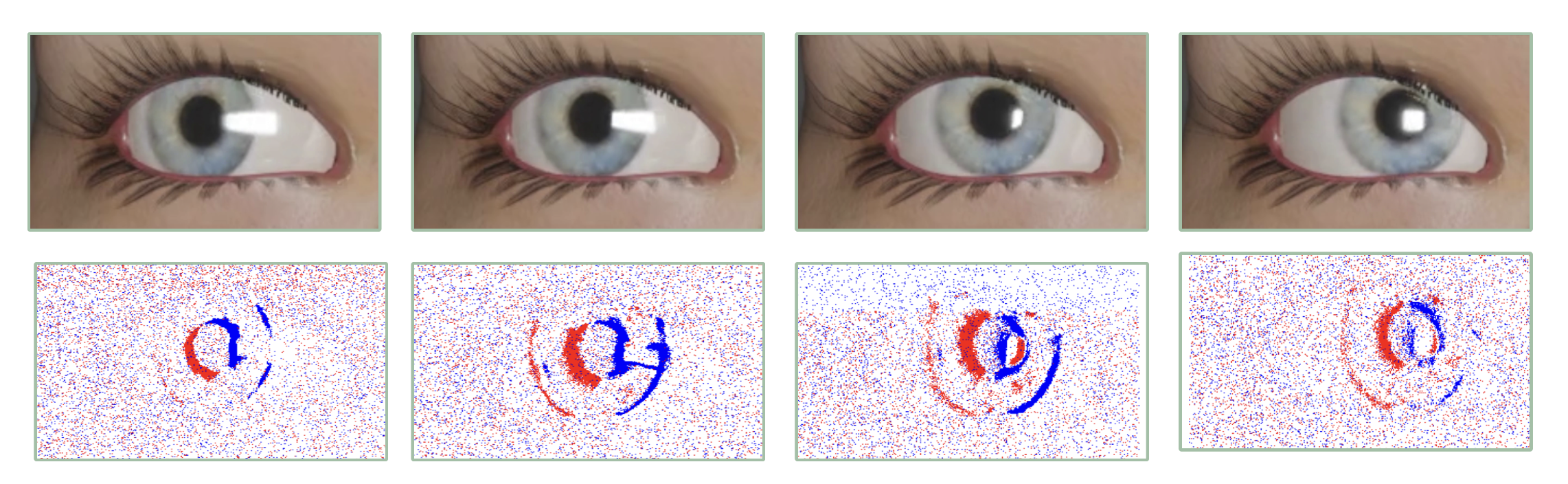}
    \caption{Visualization of a Blender-rendered saccade sequence in  frames (top row) and corresponding event stream frame representation visualized as frames (bottom row).}
    \label{fig:sac_frames}
\end{figure}

A sequence of RGB frames for a saccade and its corresponding event stream in a frame representation is illustrated in~\Cref{fig:sac_frames}.

%% file: sec/4_methodology.tex
\section{Methodology}
\label{sec:methods}
In this section, we discuss the representation of events used followed by a concise description of the network architectures employed.

\subsection{SNN Representation}
The choice of event representation is heavily dependent on the algorithms to be trained with. For SNNs, the ideal representation  is one that preserves temporal precision while remaining sparse. Common representation used in SNN-based literature include time surface representation~\cite{liang2024event}, voxel cube~\cite{cordone2022object}, binary spikes~\cite{shariff2024spiking}, etc~\cite{ruckauer2020event}. We adopt a binary spike-based representation in which each event is directly converted into a binary spike. This preserves temporal precision, minimizes redundancy while being naturally aligned with the event-driven computational paradigm of SNNs. Here, $k$ denotes the discrete time-step (temporal bin) obtained by quantizing the event timestamp $t_i$. The binary spike tensor $S(x,y,k,p)$ is defined as:

\begin{equation}
    S(x,y,k,p) =
\begin{cases}
1, & \text{if } x=x_i,\; y=y_i,\; k=k_i,\; p=p_i, \\
0, & \text{otherwise}.
\end{cases}
\end{equation}

Binary spike trains provides SNNs flexibility in how they exploit neural coding allowing them to leverage temporal coding schemes such as rate and latency coding. Once events are converted into spike trains, their interpretation within an SNN depends on the selected neural coding strategy. Common coding schemes are categorized into rate and temporal coding~\cite{nunes2022spiking}. Rate coding represents information through average neuronal firing frequency over a temporal window offering robustness to noise and training variability. For a neuron receiving a spike train 
$\{t_j\}_{j=1}^{N}$ within a temporal window of length $T$, 
the rate-coded firing intensity is given by:
\begin{equation}
    r = \frac{1}{T} \sum_{j=1}^{N} \mathbf{1}(t_j \in [0, T]),
\end{equation}
where $r$ denotes the firing rate in spikes per unit time.

Alternatively, if $S(t)$ is the binary spike signal:
\begin{equation}
    S(t) = 
\begin{cases}
1, & \text{if a spike occurs at time } t, \\
0, & \text{otherwise},
\end{cases}
\end{equation}

then the rate-coded value is
\begin{equation}
    r = \frac{1}{T} \int_{0}^{T} S(t)\, dt.
\end{equation}

Temporal and latency-based encoding strategies, by contrast, leverage precise spike timing and are often regarded as well suited for tasks that depend on fine-grained temporal information. However, these approaches tend to be highly sensitive to timing noise and fluctuations in event rates, which can hinder training stability. In this work, Rate coding is adopted to promote a more robust convergence, enhance training stability, mitigate variability arising from event‑rate fluctuations, and provide more predictable optimization dynamics, particularly for lightweight architectures~\cite{guo2021neural}. The resulting spike trains derived from EC data and encoded via rate coding are used to train the SNN models discussed in~\Cref{sec:SNNs}

\subsection{Spiking Neural Networks Adapted}
\label{sec:SNNs}
The operations of SNNs are governed by the dynamics of spike-based communication, where information is transmitted through discrete action potentials rather than continuous values. Each neuron integrates incoming post-synaptic currents, applies temporal decay to its membrane potential, and generates a spike once the potential crosses a defined threshold. Neuron models defines how inputs are integrated and transformed into spikes, synaptic dynamics determine how signals propagate across layers, and the encoding scheme specifies how external stimuli are represented as spike trains. Formally, the membrane potential \(V(t)\) of a leaky integrate-and-fire (LIF) neuron, the most common neuron, evolves according to:

\begin{equation}
  \tau_m \frac{dV(t)}{dt} = -V(t) + R \cdot I(t)
\end{equation}

where \(\tau_m\) is the membrane time constant, \(R\) is the membrane resistance, and \(I(t)\) denotes the synaptic input current.  
A spike is emitted when

\begin{equation}
    V(t) \geq V_{\text{th}}
\end{equation}

We adopt the current-based leaky integrate-and-fire (CUBA LIF) neuron model~\cite{dampfhoffer2022investigating}. This formulation captures the essential biophysical processes of membrane potential evolution by modeling synaptic input as current injection, which decays exponentially over time. The membrane potential itself undergoes leakage, ensuring stability, and spikes are emitted once the threshold is reached. The  model operates by accumulating input spikes over time, producing an output spike whenever the membrane potential crosses a predefined threshold~\cite{shariff2024spiking}. 
Unlike standard LIF models that directly integrate input into the membrane potential, Cuba-LIF separates current accumulation and voltage update, enabling more stable learning and better temporal representation. The model emits spikes when the membrane potential crosses a threshold, and resets accordingly to maintain temporal sparsity. The behavior of the Cuba-LIF neuron is governed by the following equations:

\paragraph{Synaptic Current Update:}
\begin{equation}
i[t] = \alpha \cdot i[t-1] + x[t] \end{equation} 
where $i[t]i[t]$ is the synaptic current at time tt, $x[t]x[t]$ is the input spike, and $\alpha $controls the decay rate of the current.

\paragraph{Membrane Potential Update:}
\begin{equation} 
y[t] = \beta \cdot y[t-1] + (1 - \beta) \cdot i[t] - \vartheta \cdot s[t-1] 
\end{equation} 
Here, y[t]y[t] is the membrane potential, $\beta$ is the leak rate, and the subtraction term ensures reset upon spiking.

\paragraph{Spike Emission:}
\begin{equation} s[t] = H(y[t] - \vartheta)
\end{equation}

where $ H(\cdot)$ is the Heaviside step function, and $\vartheta$ is the firing threshold.  This formulation allows Cuba-LIF neurons to maintain richer temporal dynamics and supports more effective learning in deep SNN architectures.

\subsubsection{Dense Spiking Neural Network Architecture (DenseSNN)}

We begin with a baseline model developed using the Lava framework~\cite{lava-nc}, specifically tailored for the N-MNIST dataset~\cite{orchard2015converting}. The architecture consists of three fully connected layers of CUBA neurons; An input layer receiving spike tensors of spatial resolution of 260 $\times$ $360 \times 2$. This is followed by 2 Dense layers with $512$ neurons each and finally passed to the Output layer containing 2 neurons for our binary classification task of fixation and saccades. The neuron parameters  used consistently across layers is a threshold of $1.25$, synaptic current decay of 0.25, Membrane voltage decay of $0.03$ and a Surrogate gradient with $slope = 3$ and  $width = 0.03$. Training utilises surrogate gradient descent with fixed internal neuron parameters $(requires\_grad=False)$. Delay mechanisms and weight normalization are enabled to enhance temporal learning and stability. In addition to spike-based outputs, the model records average spiking activity at each layer, which provides insights into neural responsiveness and network sparsity.

\begin{table}[hbt!]
\centering
\begin{tabular}{|c|c|c|c|}
\hline
\textbf{SNN Layer} & \textbf{$c_{in}$} & \textbf{$c_{out}$} & \textbf{Parameters} \\
\hline
Dense1 & $2 \times 260 \times 360$ & 256 & 47,954,176 \\
\hline
Dense2 & 256 & 128 & 32,896 \\
\hline
Dense3 & 128 & 2 & 258 \\
\hline
\end{tabular}
\caption{Configuration of Baseline Feedforward SNN}
\label{dense_architecture}
\end{table}

The architecture in ~\Cref{dense_architecture} highlights the enormous parameter count of the baseline feedforward SNN. The first dense layer alone requires over $47 million$ parameters making it computationally expensive and prone to overfitting. The subsequent layers even though much smaller add additional parameters resulting in a total of 48 million parameters.

\subsubsection{Convolutional Spiking Neural Network Architecture (ConvSNN)}
Building upon the baseline, we introduce an extended model that integrates convolutional layers to capture spatial hierarchies within the input data. The architecture comprises three spiking convolutional layers followed by fully connected layers for classification. A pooling layer with stride two reduces temporal noise and spatial redundancy, after which two convolutional blocks are applied with kernel size 5 and channel configurations of 2 and 8. A second pooling operation further downsamples the feature maps, aiding translation invariance.

The resulting representation ($2 \times 61 \times 86$) is flattened and passed through two fully connected layers of 512 neurons each before reaching the output layer with 2 neurons representing fixation and saccade classes. Consistent with DenseSNN, CUBA-LIF neuron parameters are applied across all layers, with gradient training enabled $(\texttt{requires\_grad=True})$. Dropout $(p=0.05)$, delay connections, and weight normalization are incorporated in the dense layers to enhance generalization and stabilize temporal representation.

\begin{table}[hbt!]
\centering
\begin{adjustbox}{max width=\columnwidth}
\renewcommand{\arraystretch}{1.0}
\begin{small}
\begin{tabular}{|c|c|c|c|}
\hline
\textbf{Layer} & \textbf{$c_{in}$} & \textbf{$c_{out}$} & \textbf{Parameters} \\
\hline
Conv1 & $2$ & $8$, $k=5$ & $2 \cdot 8 \cdot 5^2 = 400$ \\
\hline
Pool1 & $8$ & $8$ (stride=2) & 0 \\
\hline
Conv2 & $8$ & $8$, $k=5$ & $8 \cdot 8 \cdot 5^2 = 1600$ \\
\hline
Pool2 & $8$ & $8$ (stride=2) & 0 \\
\hline
Conv3 & $8$ & $2$, $k=5$ & $8 \cdot 2 \cdot 5^2 = 400$ \\
\hline
Pool3 & $2$ & $2$ (stride=2) & 0 \\
\hline
Flatten & $2 \times 27 \times 45$ & $2436$ & 0 \\
\hline
Dense1 & $2436$ & $512$ & $2436 \times 512 = 1\,247\,232$ \\
\hline
Recurrent & $512$ & $256$ & $512 \times 256 = 131\,072$ \\
\hline
Dense2 & $256$ & $2$ & $256 \times 2 = 512$ \\
\hline
\textbf{Total Parameters} &  &  & 
$\mathbf{1,281,216}$ \\
\hline
\end{tabular}
\end{small}
\end{adjustbox}
\caption{Configuration of Extended SNN with convolutional layers}
\label{conv_snn_architecture}
\end{table}

%% file: sec/5_experimentation.tex
\section{Experimentation}


\subsection{Dataset curation}

The synthetic dataset generated using Blender, as detailed in~\Cref{sec:dataset}, comprises four videos ranging from 0.1 to 6 seconds in duration, with frame counts between 250 and 10,000. Each video was segmented into left and right eye sequences, followed by cropping into near-eye regions to isolate pupil-centered activity. Annotation was guided by the taxonomy and evaluation framework proposed by Salvucci et al.~\cite{salvucci2000identifying}, which defines fixations as periods of spatially stable pupil position lasting at least 20 ms, typically associated with visual attention. Saccades were identified as rapid transitions between such fixations.

Manual annotation was performed on the left-eye sequences, classifying each frame into either fixation or saccade categories. Right-eye annotations were inferred by aligning frame indices with the annotated left eye, leveraging the synchronous nature of binocular eye movements. This procedure resulted in 1,151 fixation sequences and 1,045 saccade sequences (to be verified), with each frame cropped to a spatial resolution of  ranging from $186 \times 124 \times $ to $260 \times 360$ pixels.

For downstream evaluation of model robustness, the annotated dataset was employed for both pretraining and finetuning of the proposed architectures. Specifically, the DenseSNN model was pretrained on all 1,000 synthetic samples and initially evaluated on 30\% of real event data from the EV-Eye public benchmark. Subsequent finetuning was performed using 20\% and 50\% subsets of the real dataset, with evaluation consistently conducted on the same held-out test set.

\subsection{Implementation details}
The training process was implemented using the Lava Deep Learning SLAYER framework, which provides modular support for training SNNs~\cite{shrestha2018slayer}. Lava provides a flexible, modular system and  backpropagation-compatible framework with graded spikes, enabling differentiable learning. Given that saccades typically occur within durations of $20–200$ms, the simulation was discretized at 1ms resolution, and temporal windows were aligned with these ranges to ensure biologically plausible training. Training was conducted on  an NVIDIA GeForce RTX 2080 Ti GPU  for 100 epochs, a batch size of 8 and an initial learning rate of $0.01$. Surrogate gradients were employed to approximate the non-differentiable spike function, allowing effective backpropagation through time. with the Adam optimizer configured with a learning rate of 0.01 and weight decay of 0.0001 to promote stable convergence.

The network was trained using the SpikeRate loss, minimizing the squared error between observed firing rates and target rates $(rtrue,rfalser_{\text{true}}, r_{\text{false}})$ derived from one‑hot labels.

\begin{equation}
    \mathcal{L}_{\text{SpikeRate}} = \tfrac{1}{2} \sum_{c=1}^{C} \left( \tfrac{1}{T} \sum_{t=1}^{T} s_c[t] - \big(r_{\text{true}} y_c + r_{\text{false}} (1-y_c)\big) \right)^2
\end{equation}

\subsection{Evaluation Metrics}
Other evaluation metrics used are accuracy, precision, recall and F1 Score, formulated as:
\begin{equation}
    \text{Accuracy} = \frac{TP + TN}{TP + TN + FP + FN}
\end{equation}
\begin{equation}
    \text{Precision} = \frac{TP}{TP + FP}
\end{equation}
\begin{equation}
    \text{Recall} = \frac{TP}{TP + FN}
\end{equation}
\begin{equation}
    F1 = 2 \cdot \frac{\text{Precision} \cdot \text{Recall}}{\text{Precision} + \text{Recall}}
\end{equation}

\noindent where $TP$, $TN$, $FP$, and $FN$ denote true positives, true negatives, false positives, and false negatives, respectively. These metrics provide a comprehensive assessment of classification performance, particularly under class imbalance.

%% file: sec/6_results.tex
\section{Results }
\subsection{Experiment 1: Benchmarking synsacc dataset}
~\Cref{tab:densesnnvsconvsnn} reports the classification performance of the proposed SNN architectures under identical training conditions.
The DenseSNN, which flattens the input into a vector of size $2\times H\times W$ and applies fully-connected spiking layers, achieves an accuracy of $83.50\%$, compared to $70.75\%$ obtained by the ConvSNN that processes the input as a 2-channel $H \times W $  through convolutional layers. DenseSNN also shows lower loss (0.0797 vs. 0.0934) and better precision (0.8175 vs. 0.7081), recall (0.7850 vs. 0.7075), and F1 score (0.7794 vs. 0.7073).
These results indicate that, for the dataset and input representation used in this experiment, the fully-connected DenseSNN significantly outperforms the convolutional architecture across all metrics. The observed difference highlights the impact of network topology and input encoding on the performance of SNNs.

\begin{table}[ht]
\centering
\begin{adjustbox}{max width=\columnwidth}
\renewcommand{\arraystretch}{1.0}
\begin{small}
\begin{tabular}{|l|c|c|c|c|c|}
\hline
\textbf{Model} & \textbf{Accuracy} & \textbf{Loss} & \textbf{Precision} & \textbf{Recall} & \textbf{F1 Score} \\
\hline
DenseSNN & \textbf{0.8350} & \textbf{0.0797} & \textbf{0.8175} & \textbf{0.7850} & \textbf{0.7794} \\
\hline
ConvSNN  & 0.7075 & 0.0934 & 0.7081 & 0.7075 & 0.7073 \\
\hline
\end{tabular}
\end{small}
\end{adjustbox}
\caption{Performance comparison of DenseSNN and ConvSNN.}
\label{tab:densesnnvsconvsnn}
\end{table}

\subsection{Experiment 2: Temporal Resolution  Results of DenseSNN}

One of the key advantages of event cameras is their ability to capture visual information asynchronously with microsecond-level temporal precision. Unlike frame-based sensors, which produce images at fixed intervals, event cameras record brightness changes continuously, allowing the data stream to be partitioned into arbitrary temporal windows without losing motion fidelity. This enables flexible temporal slicing, i.e., reconstructing inputs at different temporal resolutions while preserving the underlying spatio-temporal structure required for learning.

Building on the  performance of DenseSNN in Experiment 1, we evaluated the model across time slices ranging from 200 ms down to 8 ms, training and testing independently for each temporal setting. ~\Cref{tab:temporal_performance} summarises the results. DenseSNN achieves its highest accuracy at 200 ms (92.25\%), indicating that larger integration windows provide richer temporal context. Performance remains stable down to 50–80 ms, suggesting the network effectively exploits the temporal redundancy in event streams. Accuracy declines more noticeably below 33 ms due to sparser input, but the model maintains reasonable performance even at 10–20 ms.

\begin{table}[ht]
\centering
\begin{adjustbox}{max width=\columnwidth}
\renewcommand{\arraystretch}{1.0}
\begin{small}
\begin{tabular}{|l|c|c|c|c|c|}
\hline
\textbf{TS (ms)}   & \textbf{Accuracy} & \textbf{Loss} & \textbf{Precision} & \textbf{Recall} & \textbf{F1 Score} \\
\hline
\textbf{200} & \textbf{0.9225} &\textbf{ 0.0371} & \textbf{0.9256} & \textbf{0.9225} & 0.9224 \\
\hline
150 & 0.8700 & 0.0548 & 0.8774 & 0.8694 & 0.8694 \\
\hline
100 & 0.8675 & 0.0634 & 0.8791 & 0.8675 & 0.8665 \\
\hline
80 & 0.8575 & 0.0626 & 0.8766 & 0.8575 & 0.8557 \\
\hline
50 & 0.8575 & 0.0628 & 0.8716 & 0.8575 & 0.8561 \\
\hline
33 & 0.7850 & 0.0797 & 0.8175 & 0.7850 & 0.7794 \\
\hline
20 & 0.7050 & 0.0926 & 0.7029 & 0.7000 & 0.6980 \\
\hline
10 & 0.6850 & 0.1021 & 0.7056 & 0.6825 & 0.6733 \\
\hline
8 & 0.8675 & 0.1140 & 0.6944 & 0.6625 & 0.6481 \\
\hline
\end{tabular}
\end{small}
\end{adjustbox}
\caption{Performance of DenseSNN across varying temporal resolutions on the EV-Eye dataset.}
\label{tab:temporal_performance}
\end{table}

Varying the temporal resolution introduces a natural trade-off between accuracy, responsiveness, and computational efficiency. Larger integration windows (e.g., 150–200 ms) provide denser, more informative event representations, leading to higher accuracy but lower effective frame rates. In contrast, shorter windows (e.g., 10–33 ms) significantly increase the effective frame rate, allowing the system to respond faster and process more temporal samples per second. While performance decreases slightly due to sparser input, each forward pass becomes computationally cheaper, enabling faster inference and lower power consumption.

This balance highlights a unique strength of event-based perception: temporal resolution can be tuned depending on the operational requirements. If high accuracy is critical, larger windows provide rich temporal context. If real-time responsiveness or computational efficiency is prioritized-such as in embedded driver-monitoring systems-shorter windows allow DenseSNN to run at very high effective FPS with reduced complexity. Overall, DenseSNN demonstrates robust performance across this spectrum, making it suitable for flexible, resource-aware event-driven applications.

\subsection{Experiment 3: Real-event generalization}

To evaluate how well the dense networks trained on synthetic Blender-generated event streams transfer to real neuromorphic data, we conducted a series of fine-tuning experiments using the EV-Eye dataset. As previously noted, EV-Eye is, to the best of the authors’ knowledge, the only publicly available dataset that provides raw, uncompressed event streams of eye-movement behaviour, including saccades and fixations, collected from five subjects over several minutes of recording.

The model trained purely on synthetic event streams at 33ms  was first evaluated directly on real data (“zero-shot”), achieving approximately 71\% accuracy, indicating that the synthetic training already provides a meaningful initialization but still leaves a domain gap. To quantify how much real data is required to close this gap, we fine-tuned the network using 20\% and 50\% of the EV-Eye dataset.

As shown in~\Cref{tab:real_events}, fine-tuning with only 20\% of real events resulted in a substantial improvement, raising accuracy to 86.33\% with a balanced precision–recall profile. Increasing the fine-tuning set to 50\% produced a further enhancement, achieving 87.76\% accuracy, demonstrating that the synthetic-to-real transfer continues to scale with additional real data. These results indicate that synthetic Blender-V2E event streams provide a strong pretraining signal, and that only a relatively small amount of real event data is needed to bridge the remaining domain gap.

\begin{table}[h!]
\centering
\caption{Experiment 3: Finetuning on Real Event Data (EV-Eye)}
\begin{tabular}{|l|l|c|c|c|}
\hline
 \textbf{Finetune}  & \textbf{Accuracy} & \textbf{Precision} & \textbf{Recall} \\
\hline
 1000 Blender samples & 0.7106 & 0.6318 & 0.7383 \\
20\% (200) EV-Eye &  0.8633 & 0.8646 & 0.8633 \\
 50\% (500)   & 0.8776 & 0.8788 & 0.8776 \\
\hline
\end{tabular}
\label{tab:real_events}
\end{table}

\subsection{Experiment 4: Computational Efficiency}
In SNNs, neurons communicate via discrete spikes which occur only when the neuron’s membrane potential crosses a threshold. Each spike triggers computation along the connected synapses, which update the postsynaptic neurons’ states. Consequently, the network only performs operations when a spike occurs, resulting in a highly sparse and efficient computation pattern. In contrast, ANNs use continuous activations for every neuron at each layer. Each activation involves a multiply-accumulate (MAC) operation to propagate the signal through the network, regardless of whether the neuron contributes significantly to the final output.

As shown in~\Cref{tab:computation}, the number of events and synaptic operations in the DenseSNN is orders of magnitude lower than the number of activations and MACs in the equivalent ANN. For example, layer-0 of the SNN generates only 19.36 events and 9,911 synaptic operations, while the ANN performs 512 activations and approximately 95 million MACs. This efficiency becomes even more pronounced in deeper layers: layer-2 of the SNN has almost no spikes (0.33 events and 0.66 synapses), whereas the ANN continues to compute all activations and MACs.

\begin{table}[ht]
\centering
\begin{adjustbox}{max width=\columnwidth}
\renewcommand{\arraystretch}{1.05}
\begin{tabular}{|l|r|r|r|r|}
\hline
\textbf{Layer} & \multicolumn{2}{c|}{\textbf{DenseSNN}} & \multicolumn{2}{c|}{\textbf{Equivalent ANN}} \\
\cline{2-5}
 & \textbf{Events} & \textbf{Synapses} & \textbf{Activations (ACs)} & \textbf{MACs} \\
\hline
layer-0 & 19.36 & 9911.36 & 512 & 95{,}846{,}400 \\
layer-1 & 17.91 & 9167.36 & 512 & 262{,}144 \\
layer-2 & 0.33  & 0.66     & 2   & 1{,}024 \\
\hline
\textbf{Total} & \textbf{37.59} & \textbf{19{,}079} & \textbf{1{,}026} & \textbf{96{,}109{,}568} \\
\hline
\end{tabular}
\end{adjustbox}
\caption{Comparison of the Computational efficiency  of the proposed SNN model at 33ms temporal resolution with equivalent ANN metrics}
\label{tab:computation}
\end{table}

By leveraging binary spike-based communication SNNs avoid redundant computations performed in ANNs, resulting in significantly lower memory access, energy consumption, and arithmetic cost. Even though the total number of neurons is comparable to the ANN, the event sparsity and synaptic sparsity in the SNN lead to  fewer computations, highlighting the efficiency of event-driven processing. This  is particularly advantageous for real-time and neuromorphic applications, where minimizing latency and power is critical. Overall, the table highlights how SNNs can achieve comparable functionality to ANNs while performing far fewer computations.
\section{Conclusion}
In this work, we present a synthetic dataset generated with Blender for the classification of saccades and fixations under controlled conditions. Experimental results across two Spiking Neural Network architectures demonstrates the robustness of the proposed approach and underscores the utility of synthetic data augmentation in solving the challenge of data scarcity. The architectures not only achieve high classification accuracy but also exhibit consistent performance across varying temporal resolutions, highlighting their stability in dynamic vision tasks. Moreover, the computational efficiency gains observed by SNNs reinforce their suitability for deployment in resource-constrained environments where rapid and reliable eye movement analysis is critical. Collectively, these findings contribute to event-based eye tracking especially for fine-grained eye movements such as saccades and fixations.

\section{Acknowledgment}
Authors would like to thank Dr. Hossein Javidnia (TCD) for kindly providing the Blender's base 3D armature model used in this work. 
This research was conducted with the financial support of Insight Research Ireland Centre for Data Analytics under grant no.$[12/RC/2289\_P2]$ in collaboration with authors supported by EPICs project (DTIF), Grant No. DT $2023 0459A$.